\documentclass[titlepage,oneside,12pt]{article}

\usepackage[tiny,rm]{titlesec}
\newpagestyle{trbstyle}{
	\sethead{Narayanan Edakunni, Aditi Raghunathan, Abhishek Tripathi, John Handley, Frederic Roulland}{}{\thepage}
}
\titleformat{\section}{\bfseries}{}{0pt}{\uppercase}
\titlespacing*{\section}{0pt}{12pt}{*0}
\titleformat{\subsection}{\bfseries}{}{0pt}{}
\titlespacing*{\subsection}{0pt}{12pt}{*0}
\titleformat{\subsubsection}{\itshape}{}{0pt}{}
\titlespacing*{\subsubsection}{0pt}{12pt}{*0}

\usepackage{enumitem}
\setlist[1]{labelindent=0.5in,leftmargin=*}
\setlist[2]{labelindent=0in,leftmargin=*}

\usepackage{ccaption}
\usepackage{amsmath}
\makeatletter
\renewcommand{\fnum@figure}{\textbf{FIGURE~\thefigure} }
\renewcommand{\fnum@table}{\textbf{TABLE~\thetable} }
\makeatother
\captiontitlefont{\bfseries \boldmath}
\captiondelim{\;}

\usepackage[sort,numbers]{natbib}
	\newcommand{\trbcite}[1] {({\it \citenum{#1}})}
\setcitestyle{round}

\usepackage[pagewise]{lineno}

\usepackage{totcount}
	\regtotcounter{table} 	
	\regtotcounter{figure} 	

\usepackage{mathptmx}

\usepackage[T1]{fontenc}
\usepackage{textcomp}

\usepackage{graphicx}
\usepackage{url}
\usepackage{subfigure}
\usepackage{amssymb}
\DeclareMathOperator*{\argmin}{arg\,min}

\begin{document}

	\thispagestyle{empty}

\begin{titlepage}
\begin{flushleft}

{\LARGE \bfseries Probabilistic dependency networks for prediction and diagnostics}\\[1cm]

Narayanan U. Edakunni \\
Xerox Research Centre India\\
Prestige Tech Park \\
Bangalore, India \\
narayanan.unny@xerox.com\\[1cm]

Aditi Raghunathan\\
Indian Institute of Technology Madras\\
Chennai, India\\
aditirag@cse.iitm.ac.in\\[1cm]

Abhishek Tripathi\\
Xerox Research Centre India\\
Prestige Tech Park \\
Bangalore, India \\
abishek.tripathi3@xerox.com\\[1cm]

John Handley\\
Xerox Innovation Group \\
5 800 Phillips Road, MS 128-27E \\
6 Webster, NY\\
john.handley@xerox.com\\[1cm]

Frederic Roulland\\
Xerox Research Centre Europe\\
6 Chemin de Maupertuis\\
38240 Meylan, France\\
frederic.roulland@xrce.xerox.com\\[1cm]

5827 words + \total{figure} figures + \total{table} tables = 7327

\today
\end{flushleft}
\end{titlepage}

\newpage

\thispagestyle{empty}
\section{Abstract}
Research in transportation frequently involve modelling and predicting attributes of events that occur at regular intervals. The event could be arrival of a bus at a bus stop, the volume of a traffic at a particular point, the demand at a particular bus stop etc. In this work, we propose a specific implementation of probabilistic graphical models to learn the probabilistic dependency between the events that occur in a network. A dependency graph is built from the past observed instances of the event and we use the graph to understand the causal effects of some events on others in the system. The dependency graph is also used to predict the attributes of future events and is shown to have a good prediction accuracy compared to the state of the art.

\newpage

\section{Introduction}
In the field of transportation research, frequently there arise situations where we are interested in modelling the causal relation between different events in a transport network. Example of such an event would be arrival of a bus at a bus stop during a day. Each of these events is associated with a time of occurrence of the event and these events are recurring - a particular bus service arrives at a bus stop at around the same time everyday. We can now model different attributes associated with these events like delay of arrival, demand for the bus at a bus stop, waiting time of a bus at a bus stop etc. and use this model to understand the system better and to predict attributes of future events more accurately. These events can be modelled effectively if we have the historical data pertaining to the event. 

Traditionally, data mining tools like neural networks\trbcite{NN}, support vector regression \trbcite{svr1} have been used to model events as a function of static factors like demographics, geography etc. However, these methods of modelling do not provide a holistic view of the situation. For instance, a neural network model of congestion as a function of demographics might use the static information of demographics to model and predict the demand at a bus stop at a particular time but fails to take into account the temporal aspect of demand. 

An alternative to static regression models is to use a time series\trbcite{mape,compare,gpcount} to model the event. In our running example, the demand at a particular bus stop could be modelled as a function of the time of the day making it a time series and the parameters of the time series are learned from the historical measurements of demand. However, this approach cannot capture the effect of other factors that might influence the value of the time series (like traffic accidents, breakdowns, ripple effect of congestion in other parts of the network). 

Spatio-temporal models\trbcite{stre,trflow,volatility,statespace} have been proposed in recent times to model entities that are presumed to be affected by the spatial and the temporal properties of the entity. However, most of these methods assume that there exists a continuum of values for the entity being measured. This assumption is violated in many cases; in our running example of modelling demand at a bus stop, the value of demand is valid for discrete locations in space (only points where there is a road) and cannot assumed to hold for any arbitrary point in space. The demand levels at a particular location of the route is typically dependent on discrete points in the network and is a function of the network structure rather than the spatial spread of the events. In case where the spatio-temporal models are able to model discrete spatial points, the modelling complexity is high and has not been successfully adapted to the transportation domain. In this paper, we propose {\em a generalised tool that models discrete observable events in a transportation network as random variables and builds a graphical model\trbcite{gmodel} over these variables to map the dependency between these variables}. The resulting graphical model can be used to predict attributes of future events in the system, given the values of events that have been observed till the time of making the prediction. 
\section{Dependency graph}
Events in a transportation system can be modelled as random variables and the dependency between variables can be modelled as a probabilistic relation amongst the variables. Graphical model is a framework to model and visualize the conditional dependence between random variables. The probability of a random variable conditioned on other random variables in the system can be learned from the observed values of the random variables. In a graphical model, random variables are represented as nodes of a graph with directed edges being the conditional dependence of one variable on another. The model represents the inter-dependency of different random variables and hence is a valuable framework to understand complex systems built from these variables. The other advantage of a graphical model is that when values of certain variables are observed, conditional dependencies can be used to infer the values of the dependent variables. Hence, the framework of graphical models can be used for diagnostics and a tool for prediction of unobserved variables.

In this paper, we consider a specialised instance of graphical models where the random variables are associated with discrete events that occur at certain times of the day, every day of the week. The events might(or not) be associated with a spatial component. Hence, our approach is more generic than the usual spatio-temporal methods. Events being associated with time, can be ordered according to their time of occurrence and when used to build a graphical model, forms a causal network where the probability of an event occurring later in the day is modelled as a function of a set of events that occurred earlier in the day. We can include the effect of static information like demographics, weather etc. by treating them as random variables with the difference that they do not vary with time at the same rate as the events that we model, hence they do not have a temporal component associated with it.
\section{Building a dependency graph}
\label{sec:graph}
In this section, we describe the method used to build a dependency graph from data. We can illustrate the procedure of building a model of dependency using a simple example of a bus network that consists of 7 arrival events(AE) which are identified by the sequence of numbers from $1$ to $7$ for the purposes of this example. These arrival events correspond to buses arriving at different bus stops in the network at different times of the day. An arrival event consists of a tuple of a bus stop and the scheduled time of arrival. The objective is then to model the delay at various bus stops using a dependency graph. The delays associated with these 7 arrival events are observed over a number of days for that particular bus network. 
Using this data we identify the dependency of a target AE to other AEs that preceded the target AE in time. For instance, if we observe that AEs $1,2,3$ consistently precede AE $4$ in time, we build a dependency relation that maps the delay values of AEs $1,2,3$ to the delay of AE $4$. Mathematically, we learn a probability distribution of the delay in AE $4$ conditioned on the observed delays in $1,2$ and $3$. It can be expressed mathematically as $P(a_4 | a_1,a_2,a_3)$, where $a_1 \hdots a_4$ are random variables corresponding to delay of the respective AEs.

We model dependency between AEs using a simple generalized linear relation between the outcome AE and the independent AEs. A generalized linear model is a generalization of linear regression for different noise models. The expected value of the outcome variable $a_4$ would be given by:
\begin{equation}
\mathbb{E}\left(a_4 | a_1,a_2,a_3,a_0\right)  = g(m_1a_1 + m_2a_2 + m_3a_3 + a_0),
\label{eqn:linear}
\end{equation}
where $g$ is a {\em link function} whose form depends on the choice of the linear model. When the noise is assumed to be Gaussian, the link function used would be an identity function. For a Poisson distributed outcome variable, the link would be an exponential function.
We could also enrich the model through the use of external variables in the mapping. Examples of extrinsic variables could be weather information, demographic information, the day of the week and so on. While fitting the model to the data we must try to obtain sparse models where many of the coefficients of regression ($m_1,m_2 \hdots$) are zero. The sparsity of the model ensures that only the most influential dependencies are included in the model, encourages better generalization to previously unseen data and also improves the interpretability of the model by keeping it simple. In view of this requirement, we use a lasso regression\cite{lasso} to fit a sparse model. Lasso linear regression is a procedure that encourages sparse models of linear fit, where many of the independent variables get excluded from the mapping. Sparsity of solution is encouraged by adding a regularizer term to the objective function of error of fit. More specifically, the coefficients of linear regression are obtained by solving the following quadratic program corresponding to the model given in eq.(\ref{eqn:linear}),
\begin{eqnarray}
\argmin_{m_0,m_1,m_2,m_3} \sum_i{\left(a^i_4 - g(m_1a^i_1 + m^i_2a_2 + m^i_3a_3 + m_0)\right)^2} \\
\text{subject to} \quad |m_1| + |m_2| + |m_3| \leq s.
\end{eqnarray}
where $i$ is an index over the data and $s$ is a tuning parameter. Depending on the value of $s$, the number of coefficients($m$) taking value of zero varies. Hence, $s$ can be considered to be a parameter that controls the sparsity of the regression model.

\begin{figure}
\centering
\subfigure[Dependencies between AEs]{\includegraphics[width=0.45\textwidth]{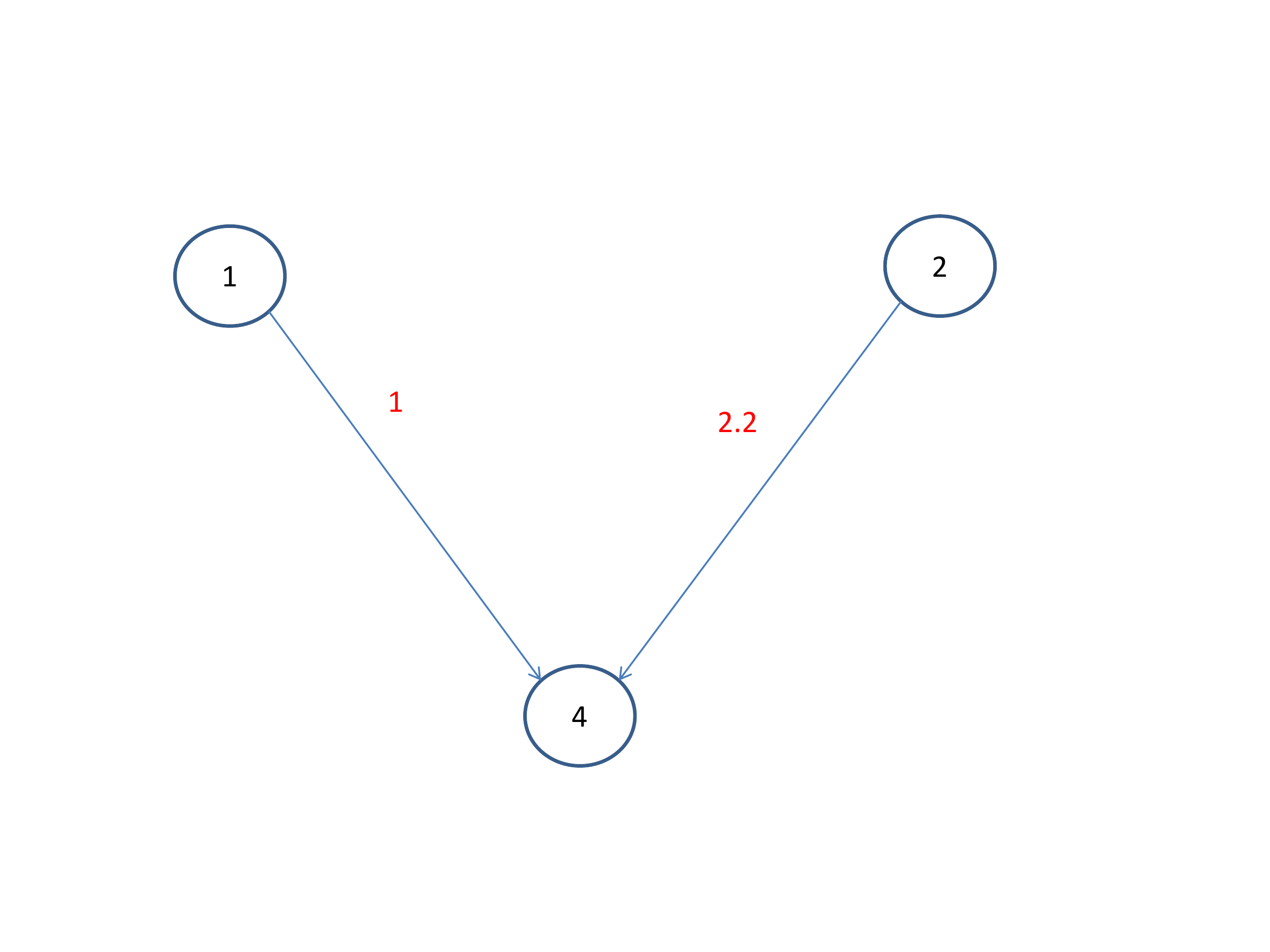}\label{fig:dependency}}
\subfigure[Directed acyclic graph of AEs]{\includegraphics[width=0.45\textwidth]{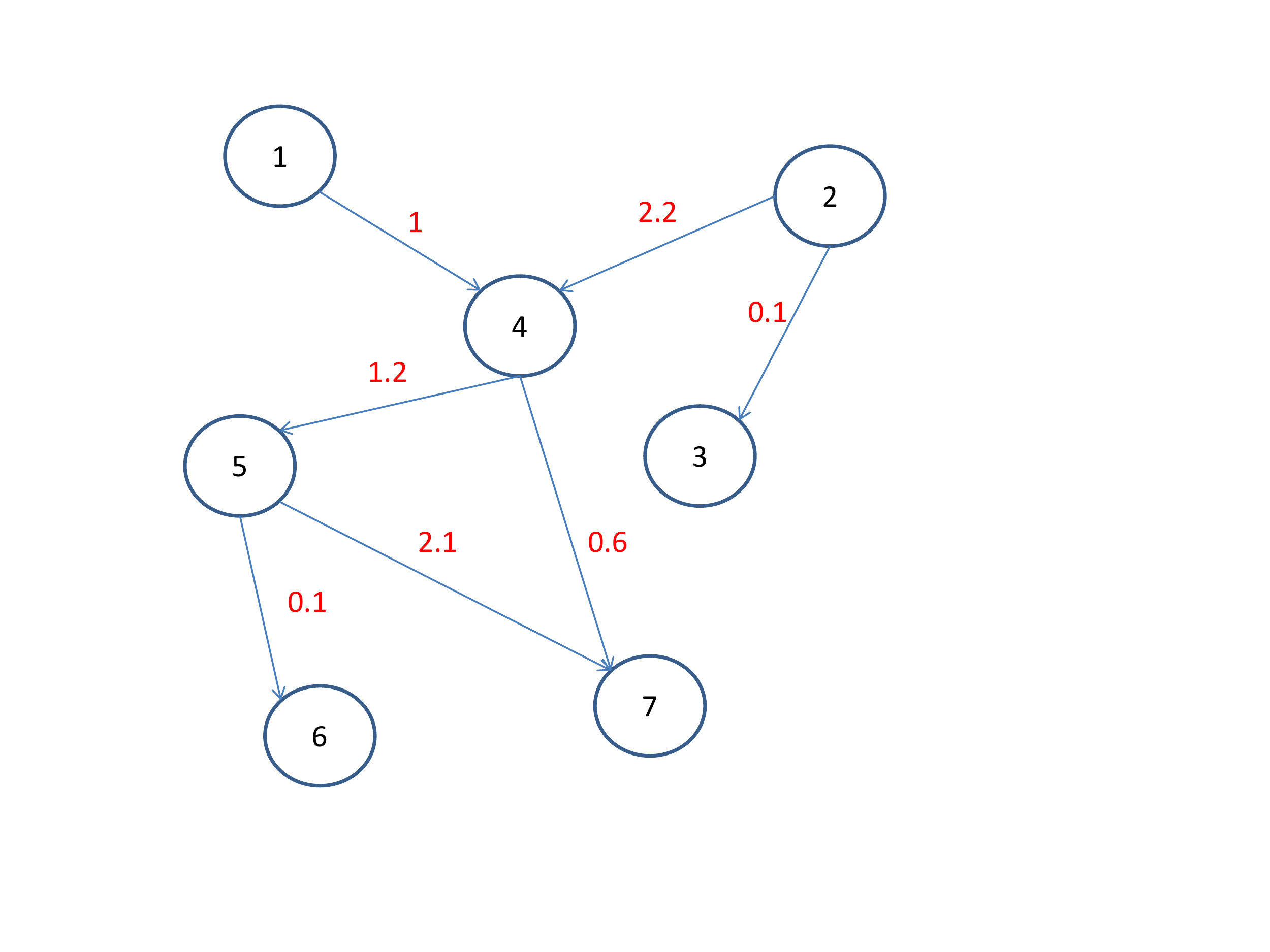}\label{fig:dag}}
\caption{Building a dependency graph}
\end{figure}

In our running example, lasso regression on the data might yield $m_3=0$, which leads to the conclusion that $a_4$ is dependent on $a_1$ and $a_2$. This dependency can be represented graphically as shown in fig.~\ref{fig:dependency}. The nodes in the graph correspond to the AEs and the directed edges represent the dependency between the AEs. The arrow on the edge indicates the direction of flow of the dependency. In fig.~\ref{fig:dependency}, $a_4$ is dependent on $a_1$ and $a_2$. The weights on the edges correspond to the strength of the dependency and is given by $m_1$ and $m_2$ of eq.(\ref{eqn:linear}).
The process of inferring dependency can be repeated for every AE in turn. The dependencies inferred can then be added to the existing nodes in the graph to obtain a directed acyclic graph(DAG) similar to one shown in fig.~\ref{fig:dag}. The graph generated from the data will necessarily be an acyclic graph because the graph is constructed in such a way that any parent node will precede a child node in time. This condition prevents any cycles in the graph. The directed acyclic graph built using the data summarises the dependencies between the different AEs. We can now use this structure to predict the delays encountered in future AEs based on the delays observed in the past.

\section{Predicting events based on the dependency graph}
\begin{figure}
\centering
\subfigure[AE 1 and 2 are observed]{\includegraphics[width=0.45\textwidth]{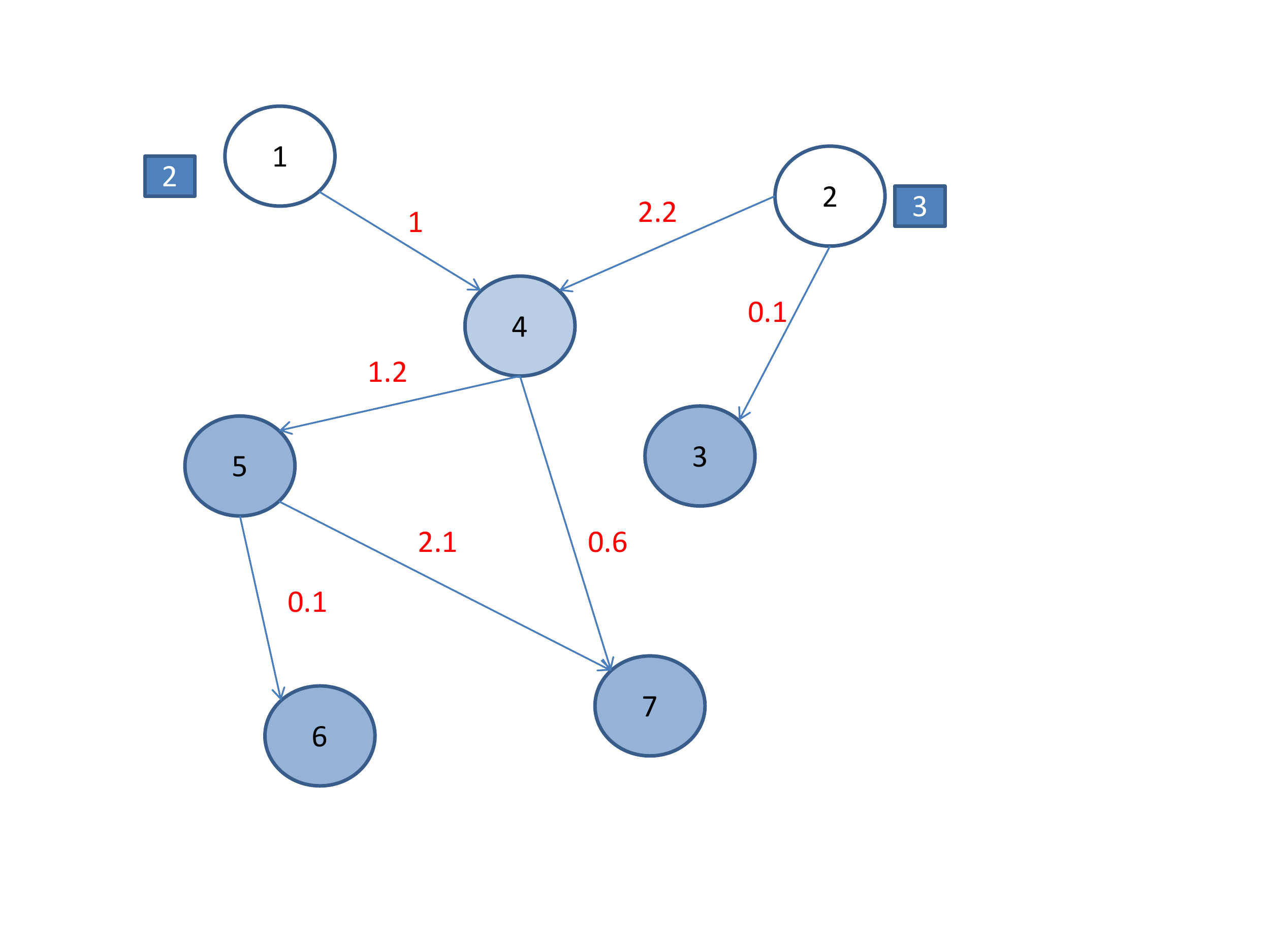}
\label{fig:predict1}}
\subfigure[Propagation of predicted delays to the descendants of the observed AEs]{\includegraphics[width=0.45\textwidth]{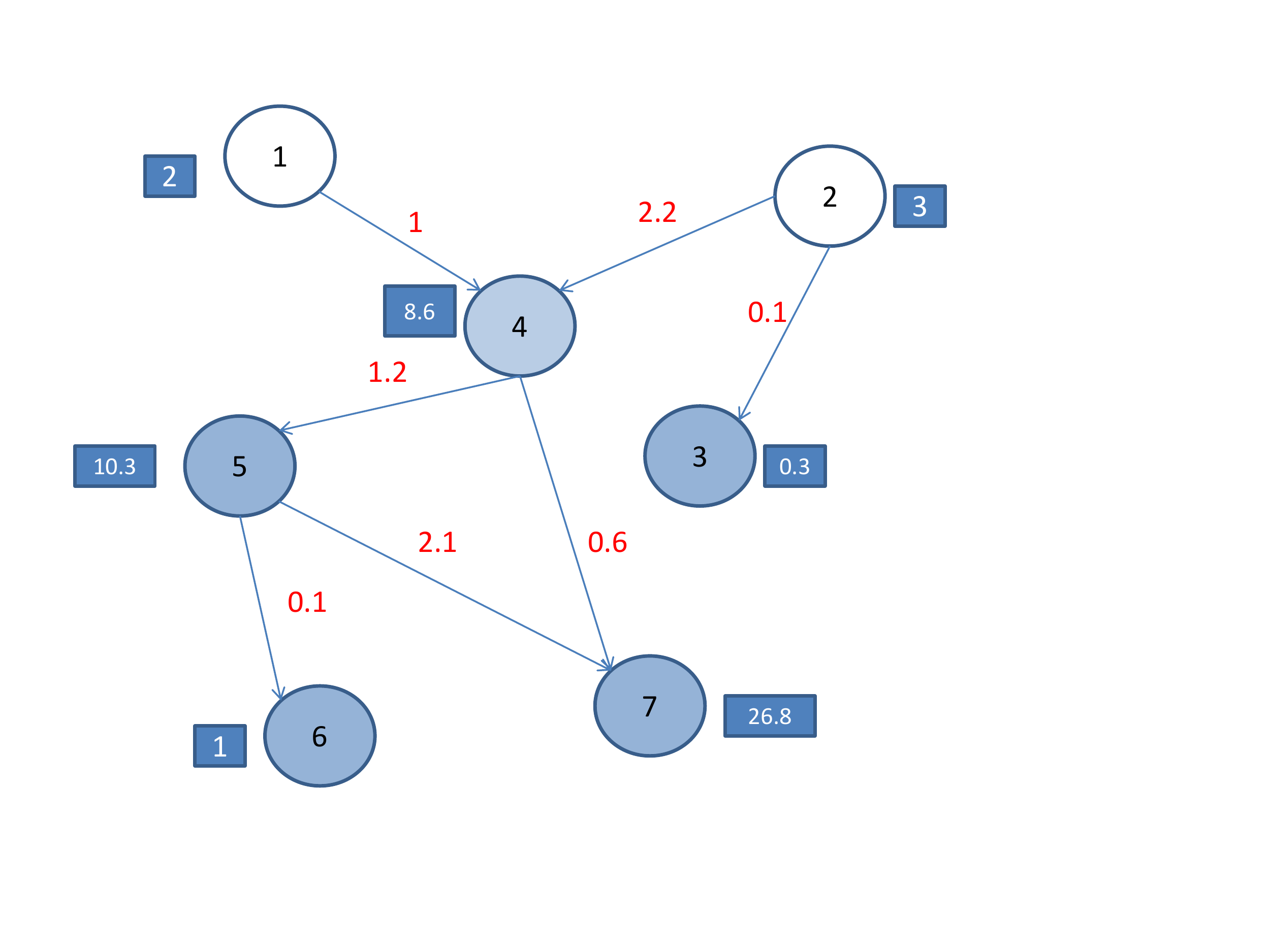}
\label{fig:predict2}}
\caption{Predicting delays using the dependency graph}
\end{figure}

When we observe the delay in an AE on a particular day we can use this observation to predict the delays of AEs that are dependent on the observed AE. The predicted delay in turn can be propagated to their children and so on. For instance, in the graph shown in fig.~\ref{fig:predict1}, AEs $1$ and $2$ are observed to have delays of $2$ and $3$ units respectively. These observed delays can then be propagated down the graph making use of the relation between the parent and the children as given by eq.(\ref{eqn:linear}). The updated graph is then shown in fig.~\ref{fig:predict2} where the filled nodes correspond to the unobserved AEs and the unfilled correspond to the AEs for which the delay has been observed. The values of the observed/predicted delays are shown beside the corresponding node.
\section{Use cases}
In this section, we illustrate applications in transportation research where a dependency graph can be used to model the relation between entities and use it for prediction. Specifically, we look at three different use cases - modelling delays in transportation networks, modelling demand in public transportation networks and modelling traffic counts on road networks. The first use case is an exemplar for use of dependency graph where the variables take values from the domain of real  numbers. Second and third use cases correspond to a scenario where random variables involved are counts and hence are values from the domain of whole numbers.

In the use cases given below, we also compare the proposed algorithm with the state of the art applicable for that particular use case. The baseline we use in each of these evaluations is the mean prediction. The mean prediction corresponds to the mean of the quantity measured over the days when the event repeats.

\subsection{Predicting delays}
A straightforward use case is to model delays in a transportation network as a dependency graph where the delays in parts of the network affect other parts of the network at later times of the day. We discuss two different instances of transportation where this technique can be applied.
\subsubsection{Predicting delays in road network}
Travel time prediction has seen a lot of research activity with methods such as neural network\cite{NN}, Kalman filter\trbcite{Kalman}, and support vector regression\trbcite{svr1,svr2} being used to solve the problem. The solutions can be broadly classified into two categories. The first solution involves dynamic tracking of the vehicle to predict the likely time of arrival at a point of interest and uses techniques like Kalman filtering \trbcite{Kalman} for this. The second class of solutions use historical information to map a set of static features(like time of day, bus stop, route information) to the delay at a particular bus stop of interest\trbcite{NN}. Both of these solutions have their own limitations. The former approach relies heavily on the tracking signal(like GPS) and is affected by noise in the signal, it also assumes a certain model of movement of the bus and hence is effective only for short range predictions. In the latter approach, models use features which are static. This results in a prediction that has a good accuracy for long range predictions but cannot adapt to changing dynamics of the delay in the transit network. A dependency graph combines the two different approaches to obtain a dynamic yet a robust prediction pattern. In this method, we make use of the delays observed at bus stops in a network during a day of service to predict the likely delays in the network, in the later part of the day. The main advantages of using this method of prediction over others are:
\begin{itemize}
\item It does not rely on tracking signals like GPS or bluetooth and hence is resistant to outages and noise.
\item It models the spatio-temporal dependency of delay at various bus stops in the network. Usually methods used to predict the delays restrict themselves to a particular route in the network and does not include the whole network in the analysis. There could be long distance(in time and space) dependence in the network that could affect delays at a particular bus stop and it would be crucial to include these in the analysis for a more accurate prediction.
\end{itemize}  
We obtained operational data from bus transportation network for some of the clients of our organisation and applied our algorithm to predict delays in the network. The method was able to predict at a level of about $10\%$ Mean Average Percentage Error(MAPE)\trbcite{mape} which measures error as a percentage of the actual delay. The data being proprietary cannot be shared publicly with the research community and hence we choose a different use case in the next section to illustrate the effectiveness of our algorithm over a publicly available dataset.

\subsubsection{Predicting delays in flight network}
Delays in flight networks are quite common and has received much attention recently from the statistics\trbcite{stat} and data mining community\trbcite{gequest} due to the availability of historical data. There have been many different approaches tried to predict the delay in flights at different airports\trbcite{flightdelay1,flightdelay2,flightdelay3,flightdelay4}. However, these approaches do not take into account delay propagation across the network. The research that is related to our approach is \trbcite{flightdelaynet} where the authors consider the effect of delays in other airports on the delay in the target airport. However, they do not build a network of delays and hence the propagation of delay predictions is not performed. 

\begin{figure}
\includegraphics[width=0.9\textwidth]{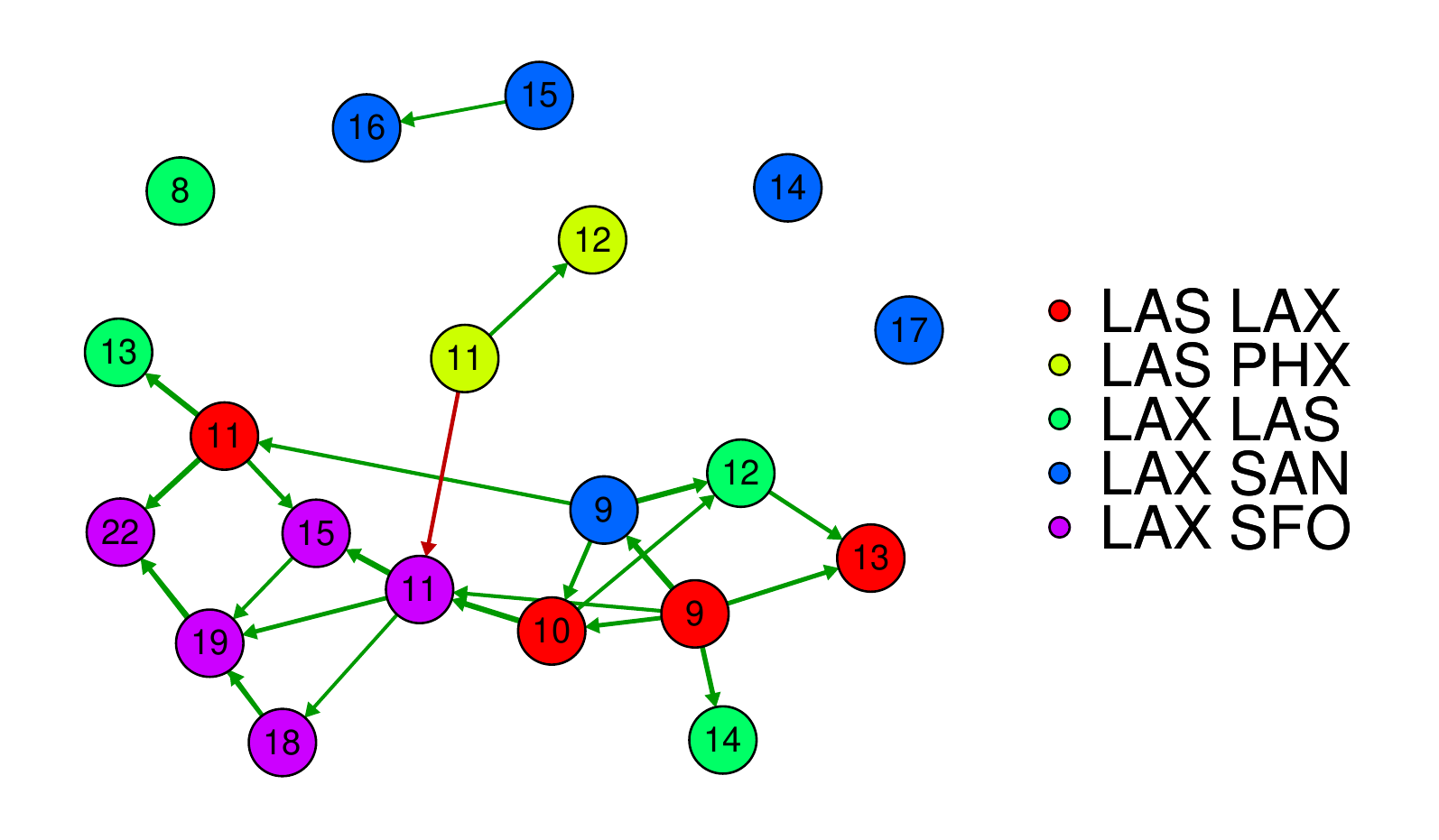}
\caption{The dependency graph relating the delays at various origin-destination links in a flight network}
\label{fig:flightdag}
\end{figure}

In our experiments, we use the data from \trbcite{stat} and arrival event as the tuple of \{Origin airport, Destination airport, scheduled time of arrival\}. The scheduled time of arrival is mapped to the hour of the day by taking the mean of the delays observed in the link during an hour. Hence, we assume that there is a pattern to the mean delay over all the flights from a particular origin to a destination at a particular hour of the day. We chose data for $285$ days of year $2008$ from \cite{stat}. We split the data into training set of $100$ consecutive days. We used this dataset to build the dependency graph and used the dependency graph to predict delays at various times of the day for the rest of the $185$ days. We used Lasso regression with a constraint on the maximum number of parent nodes set to $5$. The resulting DAG had $20$ nodes as is shown in fig.(\ref{fig:flightdag}) where the links shown in green correspond to positive values of the coefficient and red corresponds to negative weights. The intensity of the colour is proportional to the magnitude of the weight. Nodes with the same colour correspond to the same origin-destination pair and the label on the node corresponds to hour of the day. For instance, the node coloured red and labelled $9$ corresponds to the flights leaving $LAS$ and reaching $LAX$ from 9:00am to 10:00am. In the DAG shown in fig.(\ref{fig:flightdag}), we have retained only the most significant links to obtain a better visualisation. 

\begin{figure}
\includegraphics[width=0.9\textwidth]{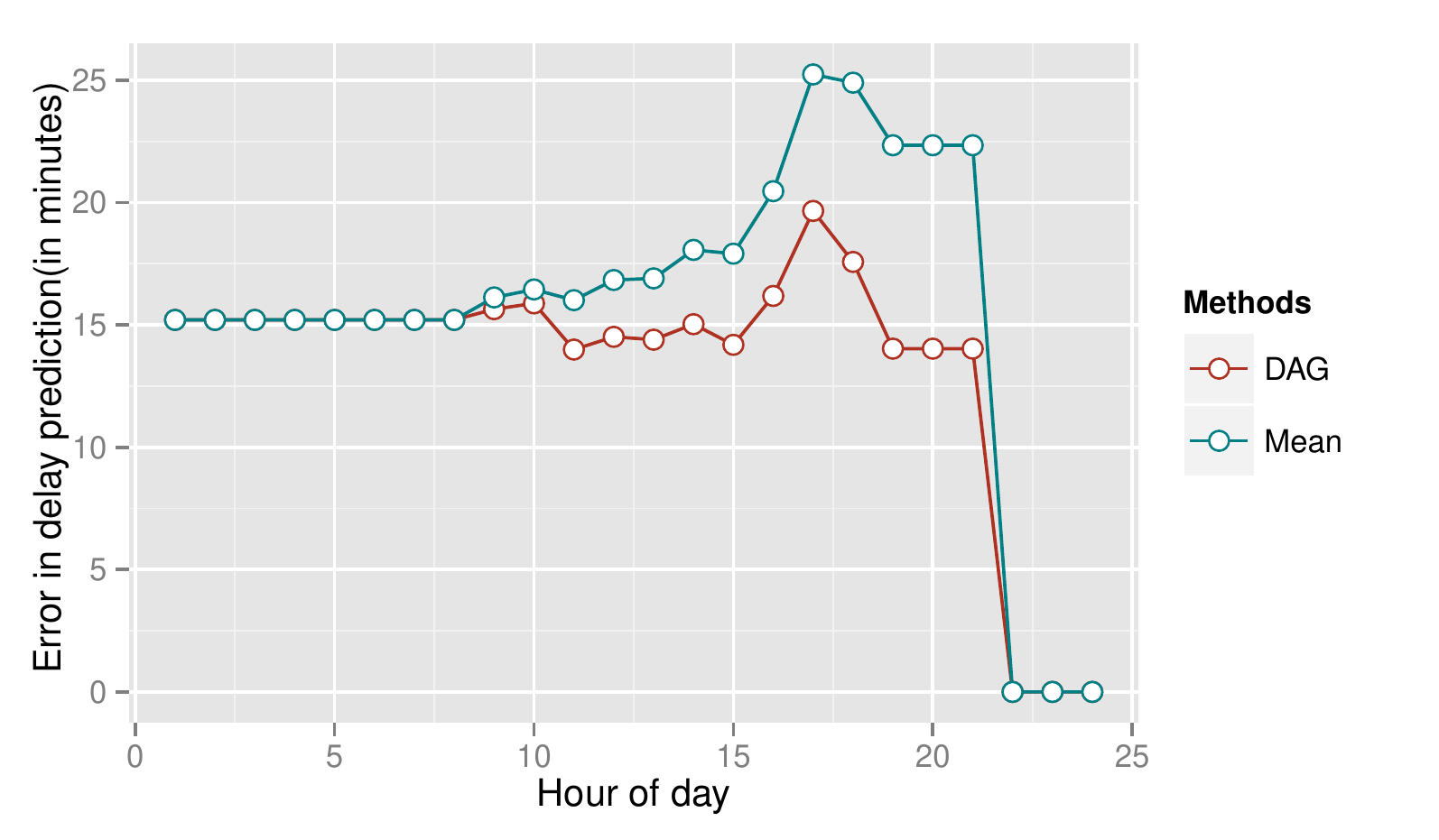}
\caption{The error in predicting the delays over various links in the flight network as a function of hour of the day.}
\label{fig:flighterror}
\end{figure}

We used the dependency graph used to predict delays on the test set and the average error in prediction as a function of hour of the day is shown in fig.(\ref{fig:flighterror}). The errors shown in fig.(\ref{fig:flighterror})were computed at each hour of the day. At each hour of the day, the delays for different origin-destination pair were computed for the rest of the day. The predictions for the past are set to the actual delays. The error in prediction is computed as the average absolute difference between predicted delays and actual delays, over the entire set of origin-destination pairs. The average error is recorded as the prediction error for the hour of the day when the prediction was made. In fig.(\ref{fig:flighterror}) we also compare our method with the baseline of predicting the mean of the delays as observed historically. We are not able to compare our method with \trbcite{flightdelaynet}, since it uses many preprocessing steps on the data which is not obvious to implement. We find that the error in prediction using our algorithm is much less than the case when a mean based prediction is used. The evolution of error illustrates the advantage of using our algorithm. At the start of the day(around 7:00am) when there is very little information about the delays, the performance of mean based prediction and our algorithm are the same. As the day progresses, delays are observed for different nodes of the DAG and the predictions obtained become more accurate and hence we see a sharp increase in performance as compared to mean based prediction as the day progresses. At the end of the day(around 22:00) when all the delays have been observed the performance of the mean based prediction and our algorithm converges.


\subsection{Predicting count data}
In this section, we consider a use case where we need to model count data using the dependency graph. We use a conditional Poisson distribution to model the count data. The dependent variable is modelled as a generalized linear model of the independent variables. The only modification we need to do the algorithm is to plug-in the appropriate link function in eq.(\ref{eqn:linear}) to model counts. The methodology remains the same as discussed in Section \ref{sec:graph}. We illustrate the effectiveness of dependency graph in two different settings.
\subsubsection{Predicting demand}
In recent times, with the advent of smart card based ticket payments information about the demand at a bus stop can be obtained in real time. The demand data obtained can then be used to predict the demand in future at different parts of the city. Advanced prediction of demand can be effectively used to fine tune the services of a transit agency and allow the authorities to take decisions in real time. Using the ticketing data obtained from clients of our organisation, it was observed that we are able to predict upto an accuracy of about $7-10\%$ of the actual demand and the prediction could be made upto $5$ hours in advance. The accuracy reported here is an average over short-term and long-term predictions. However, the data on which the experiments were conducted are proprietary and to illustrate the effectiveness of our method on publicly available datasets we conduct experiments on these datasets in the next section.

\subsubsection{Predicting traffic counts}
\begin{figure}
\centering
\subfigure[Dependency graph between the traffic volumes at different locations at different times of the day.]{\label{fig:washingtondag}\includegraphics[width=0.45\textwidth]{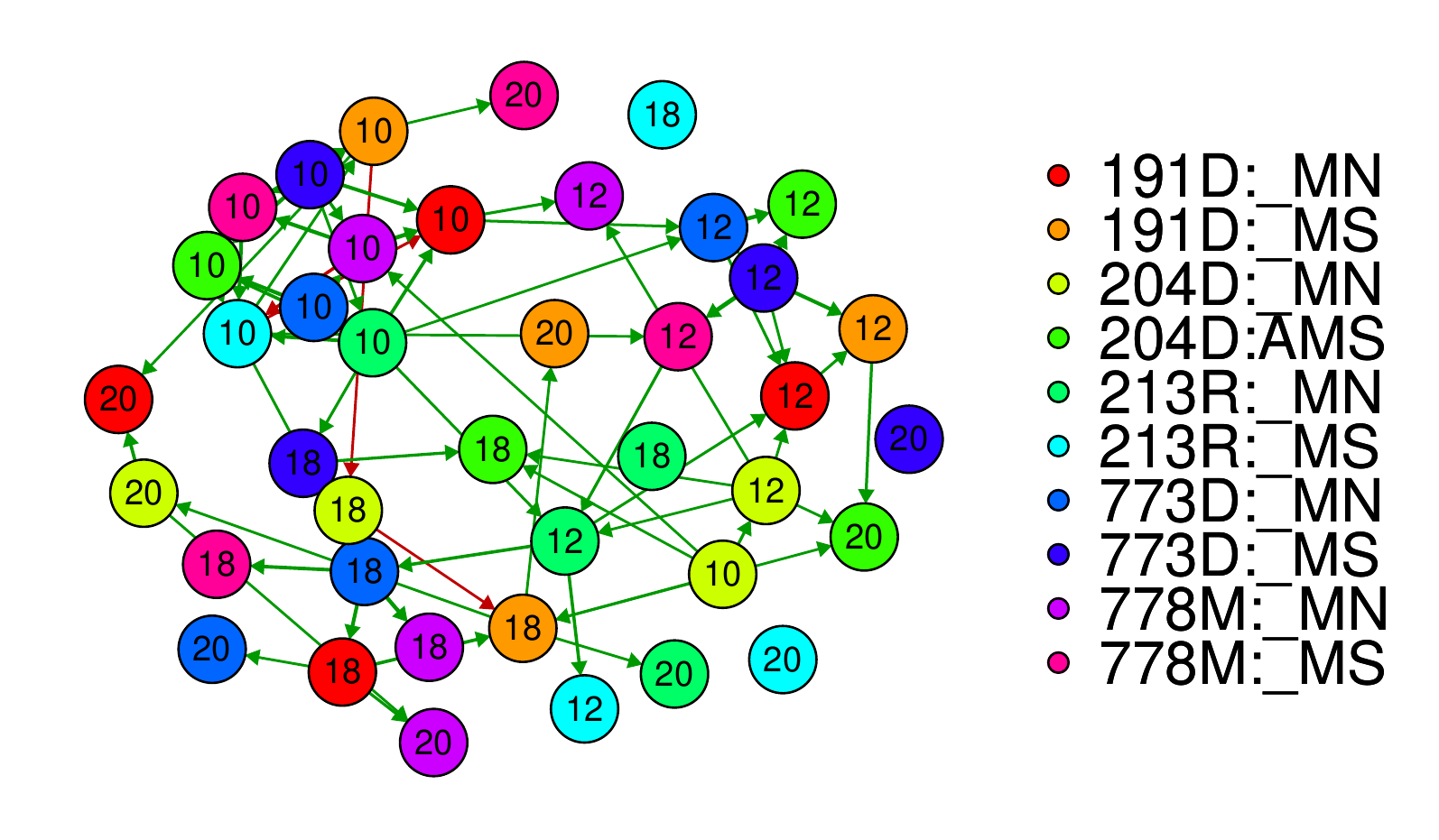}}
\subfigure[Geographical location of traffic count stations]{\includegraphics[width=0.45\textwidth]{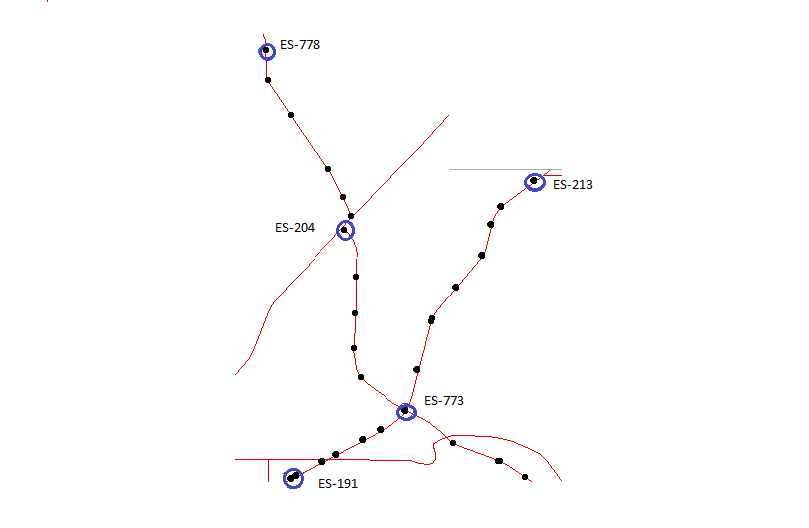}\label{fig:washingtonmap}}
\caption{Traffic counts from Washington}
\end{figure}

In this section, we look at the problem of predicting traffic counts in different parts of a road network at a particular hour of the day, given the traffic counts at other locations of the network at earlier times of the day. We collected traffic count data from \cite{TDAD} for the city of Washington. We chose five traffic count stations - $ES-773D$, $ES-204D$, $ES-778M$, $ES-213R$ and $ES-191D$ with counts collected for directions of North and South. The layout of these stations is shown in fig.(\ref{fig:washingtonmap}). The data was collected for the entire month of January in the year 2007. The counts were aggregated in hourly bins and the entities in the dependency graph was the \{Station, Direction and Hour of day\}. We use the first $20$ days to build the dependency graph and the next $10$ days to predict the aggregated counts for each hour of the day. The DAG learned as a result of the training is shown in fig.(\ref{fig:washingtondag}). The DAG shows the nodes at larger intervals of time so that we can analyse the behaviour of the macroscopic dependencies between the traffic volumes at various locations. The error in predicting the traffic counts using the dependency graph technique is compared against the mean prediction and a more sophisticated approach of ARIMA used for each of the nodes in the graph. ARIMA has been a popular method for traffic count prediction\trbcite{arima1,arima2,arima3}. Gaussian Process(GP) based time series analysis\trbcite{gpcount} has been successful recently, but the GP based prediction works well only for short-term predictions and does not yield an interpretable model. Hence, we do not consider a GP model in our comparison. The prediction error is measured as the absolute difference of the predicted count and the actual value as a percentage of the actual counts. For ARIMA, the models were learned using a {\em auto.arima} as described in \trbcite{gpcount}. The ARIMA model were trained on the training set of data and was tested on the entire day, the model was not updated with the observations of the test day. We can see that the predictions of the dependency graph is more accurate than comparable methods and even for a long term prediction the error is only about $10\%$ of the actual count.

\begin{figure}
\includegraphics[width=0.9\textwidth]{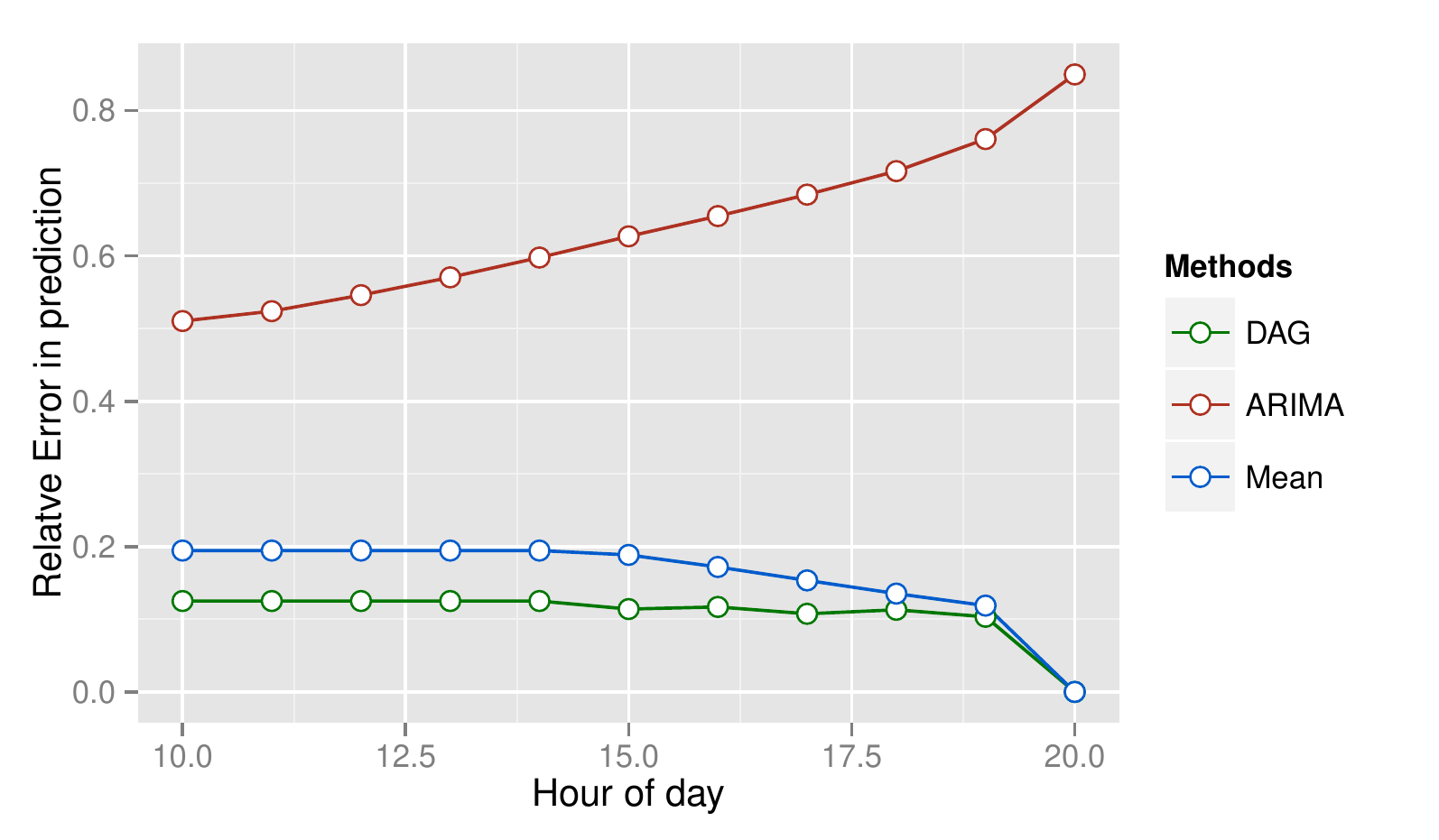}
\caption{Comparison of average error in predicting the traffic volume at different locations at different times of the day.}
\label{fig:washingtonerror}
\end{figure}

The dependency graph given in fig.(\ref{fig:washingtondag}) also provides valuable insights regarding the relation between traffic volumes at different locations. We can make the following inferences about the traffic flow in the network - 
\begin{itemize}
\item There is a lot of correspondence between overall traffic levels at any time. That is, higher traffic at one station at some time, indicates higher traffic at other stations also, at the same time(similar to what we observe daily). For example, all nodes labelled 10 or 20 are correlated.
\item The volume of north flowing traffic at 191D at around 10AM affects the traffic at 778M and 773D, at 12PM. This indicates that the north flowing traffic from 191D, usually goes towards 778M and 773D, and this is confirmed from the geographical locations of the 3 stations(and there are no edges between the south flowing traffic at 191D at 10 and traffic at 778M or 773D at 12PM)
\item Similarly, north flowing traffic at 6PM on 191D, also has edges to the north flowing traffic at 778M and 773D at 8PM, but not the south flowing traffic at either of these stations.
\item There are edges between the same station at 10AM (or 6PM) and 12PM(or 8PM) saying that the traffic at 12PM (or 8PM) is correlated to the traffic at 10AM(or 6PM), which is along the lines of common observation
\item There is a dependency between the traffic going north at 204D in the morning to the traffic going south at 204D in the evening (refer to link between light green 10 to dark green 20). 
\item Interestingly, the traffic flowing south from 213R in the evening, does not depend on the traffic in other places at earlier times of the day, which is explained by the geography, where 213R is the northern most point in the route.
\end{itemize}
The inferences listed above shows the effectiveness of the dependency graph as a diagnostic tool and can be used to study the properties of the network like bottlenecks, unintended dependencies and so on.
\section{Conclusion}
We find from our experiments over different use cases that dependency graph is a versatile tool to model the probabilistic dependency between events in a transportation network. The comparative evaluation established the efficacy of the model in predicting attributes of a event within a short time horizon as well as long time forecasts. The dependency graph can also be utilised as a diagnostic tool to visualize and understand the dynamics of a complex system. One such interpretation was given for the traffic network based on the visual examination of the DAG produced by the algorithm. However, we can use sophisticated graph analysis tools to come up with more useful insights.

\bibliographystyle{trb}
\bibliography{biblio}

\nolinenumbers

\end{document}